\documentclass[preprint,12pt]{elsarticle}

\usepackage{amssymb}

\usepackage{amsmath}

\usepackage{graphicx}

\usepackage{epstopdf}

\usepackage{subcaption}

\biboptions{sort&compress}

\usepackage{bm}

\begin{document}

\begin{frontmatter}

\title{High-precision visual navigation device calibration method
based on collimator} %% Article title

\author[label1,label2]{Shunkun Liang\fnref{fn}}

\author[label1,label2]{Dongcai Tan\fnref{fn}}

\author[label1,label2]{Banglei Guan}

\author[label1,label2]{Zhang Li}

\author[label1,label2]{Guangcheng Dai}

\author[label1,label2]{Nianpeng Pan}

\author[label3]{Liang Shen}

\author[label1,label2]{Yang Shang}

\author[label1,label2]{Qifeng Yu}

\fntext[fn]{These authors contributed equally.}

\affiliation[label1]{organization={College of Aerospace Science and Engineering, National University of Defense Technology},%Department and Organization
            city={Changsha},
            postcode={410073}, 
            state={Hunan},
            country={China}}

\affiliation[label2]{organization={Hunan Provincial Key Laboratory of Image Measurement and Vision Navigation},%Department and Organization
            city={Changsha},
            postcode={410073}, 
            state={Hunan},
            country={China}}

\affiliation[label3]{organization={North Navigation Control Technology Co., Ltd},%Department and Organization
            city={Beijing},
            postcode={100176}, 
            state={Beijing},
            country={China}}
%% Abstract
\begin{abstract}
Visual navigation devices require precise calibration to achieve high-precision localization and navigation, which includes camera and attitude calibration. To address the limitations of time-consuming camera calibration and complex attitude adjustment processes, this study presents a collimator-based calibration method and system. Based on the optical characteristics of the collimator, a single-image camera calibration algorithm is introduced. In addition, integrated with the precision adjustment mechanism of the calibration frame, a rotation transfer model between coordinate systems enables efficient attitude calibration. Experimental results demonstrate that the proposed method achieves accuracy and stability comparable to traditional multi-image calibration techniques. Specifically, the re-projection errors \( \leq 0.1463\) pixels and average attitude angle errors \(\leq 0.0586\,\text{°}\) with a standard deviation \(\leq 0.0257\,\text{°}\), demonstrating high precision and robustness.
\end{abstract}

%% Keywords

\end{frontmatter}



\section{Introduction}
\label{sec1}
%% Labels are used to cross-reference an item using \ref command.

With the rapid development of high-tech fields such as unmanned aerial vehicles (UAVs), intelligent robots, and autonomous driving, high-precision positioning and navigation have gradually become one of the key technologies in global technological competition \citep{Zhang2024, YANG2023}. In line with these mission requirements, visual navigation technology has become a research hotspot in this field due to its lightweight, high precision, robustness, and ability to achieve autonomous navigation \citep{Guo2019, Yu2025, Xu2024}. Visual navigation systems utilize cameras to capture images and combine advanced image processing and computer vision algorithms to determine location, attitude, and motion trajectories. Compared to GPS, visual navigation systems can provide accurate and reliable navigation guidance when GPS signals are limited \citep{Wang2023}. 

To enable accurate guidance of visual navigation devices, two critical challenges must be addressed: the calibration of camera parameters and the attitude of the camera relative to the reference. 

Current methods for camera calibration are divided mainly into traditional methods, which rely on known size information from calibration objects \citep{Bu2021, Zhang2000, Tsai1987, Duan2024, Liu2024}, and self-calibration methods, which utilize geometric constraints from scene information \citep{Hartley1997, Bang2019, Da2012, Yang202322, Guan2017}. Certainly, numerous researchers have also achieved three-dimensional imaging based on artificial intelligence algorithms \citep{Lei2024, Qi2024, Yu2024}. Due to their maturity and robustness, traditional methods are widely adopted for camera calibration. In the field of traditional methods, two methods have become particularly well-known for their effectiveness and precision: Zhang's chessboard calibration method \citep{Zhang2000} and Tsai's two-step method \citep{Tsai1987}. However, both require multiple captures of the calibration object from various perspectives, making the process cumbersome, time-consuming, and unsuitable for batch operations. 

Furthermore, limitations in assembly technology and potential positional shifts of the camera during prolonged use often result in misalignment between the camera coordinate system and the navigation reference coordinate system. This misalignment introduces significant deviations, adversely affecting navigation and positioning accuracy. Consequently, a precise attitude calibration is essential. Attitude calibration \citep{Bang2022, Gao2017, Fraundorfer2018, Bala2016, Li2019, Guan2023, Liu2024line, Guan2022} is typically addressed using precision instruments such as three-axis turntables \citep{Gao2017}, laser trackers \citep{ Bala2016}, total stations \citep{Li2019}, and so on. Although these tools improve measurement efficiency, they often involve complex installation procedures and may introduce cumulative errors. To streamline the calibration process, some researchers have developed integrated calibration toolboxes \citep{Francois2022, Yan2022, Salah2024}.

In response to the aforementioned challenges, we propose a novel calibration method and system for navigation devices based on a collimator. A calibration frame with a special geometric structure is proposed to facilitate rapid calibration. This frame is characterized by its low cost, ease of operation, convenience, and support of batch calibration processes. Using the optical geometry of the collimator, a single-image camera calibration method is introduced. The proposed method significantly reduces the time costs while achieving accurate calibration of the camera parameters. Furthermore, based on the geometric structure of the calibration frame, a method is designed to estimate the relative rotation between the camera and the reference, which simplifies the calibration process and reduces overall engineering costs.

\section{Visual navigation device calibration system}
This section introduces a visual navigation device calibration system. The system integrates with a collimator and a calibration frame, designed to provide stable and reliable calibration data.
\label{sec2}

%% Use \subsection commands to start a subsection.
\subsection{Design of collimator}
\label{subsec21}

\begin{figure}[htbp]
\centering
\begin{subfigure}[b]{0.55\textwidth}
    \centering
    \includegraphics[width=\textwidth]{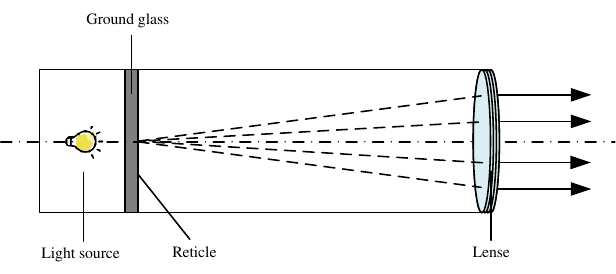}
    \caption{}
    \label{fig1a}
\end{subfigure}
\hfill
\begin{subfigure}[b]{0.33\textwidth}
    \centering
    \includegraphics[width=\textwidth]{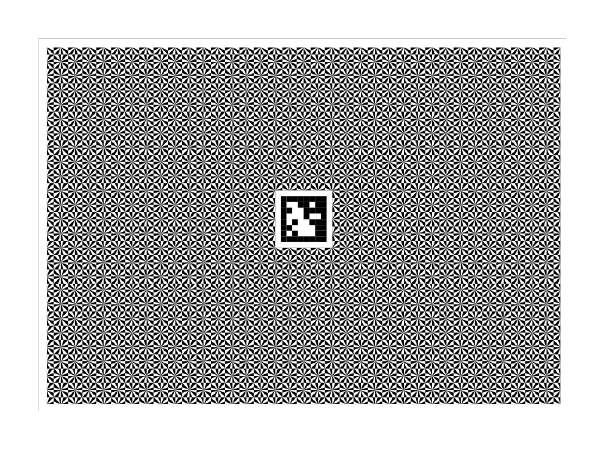}
    \caption{}
    \label{fig1b}
\end{subfigure}
\caption{Collimator design drawing. (a) Collimator structural sketch and (b) encoded marker pattern.}
\label{fig1}
\end{figure}

The collimator is a key optical instrument widely used in photoelectric detection and precision measurement systems \citep{Alexander2017, Liang2024, Liu2021}. As shown in Fig. \ref{fig1a}, the collimator consists mainly of a light source, ground glass, reticle, and lens. The light source provides stable and continuous illumination. The ground glass transforms the nonuniform distribution of the original light source into a uniform surface light source, ensuring even illumination of the reticle. The encoded marker pattern on the reticle is customized according to specific calibration requirements. The lens, a core component of the collimator, generates a parallel beam of light by refracting incident rays.

According to the principles of geometric optics, rays originating from the focal plane of a lens are refracted into a parallel beam. Using this principle, the reticle is designed with specific encoded marker patterns as calibration targets. Ray emitted from each point on the target through the lens, producing parallel beams that simulate an infinite distance. Consequently, the camera captures an image of the target through the lens, equivalent to observing the target at infinity.

A composite encoding marker pattern is customized on the reticle, as shown in Fig. \ref{fig1b}. The pattern adopts a star-type calibration pattern \citep{Thomas2020}, comprising multiple star-shaped grids and a central Apriltag pattern \citep{Edwin2011}. Unlike traditional chessboard patterns, which rely on alternating black and white squares to form corner points, the star-type pattern provides a richer diversity of gradient information, enhancing the precision of feature extraction. The Apriltag pattern facilitates the identification of feature points, enabling efficient localization and matching. This design significantly improves the accuracy and robustness of the calibration process.

\subsection{Design of calibration frame.}
\label{subsec22}

\begin{figure}[htbp] 
\centering 
\includegraphics[width=0.8\textwidth]{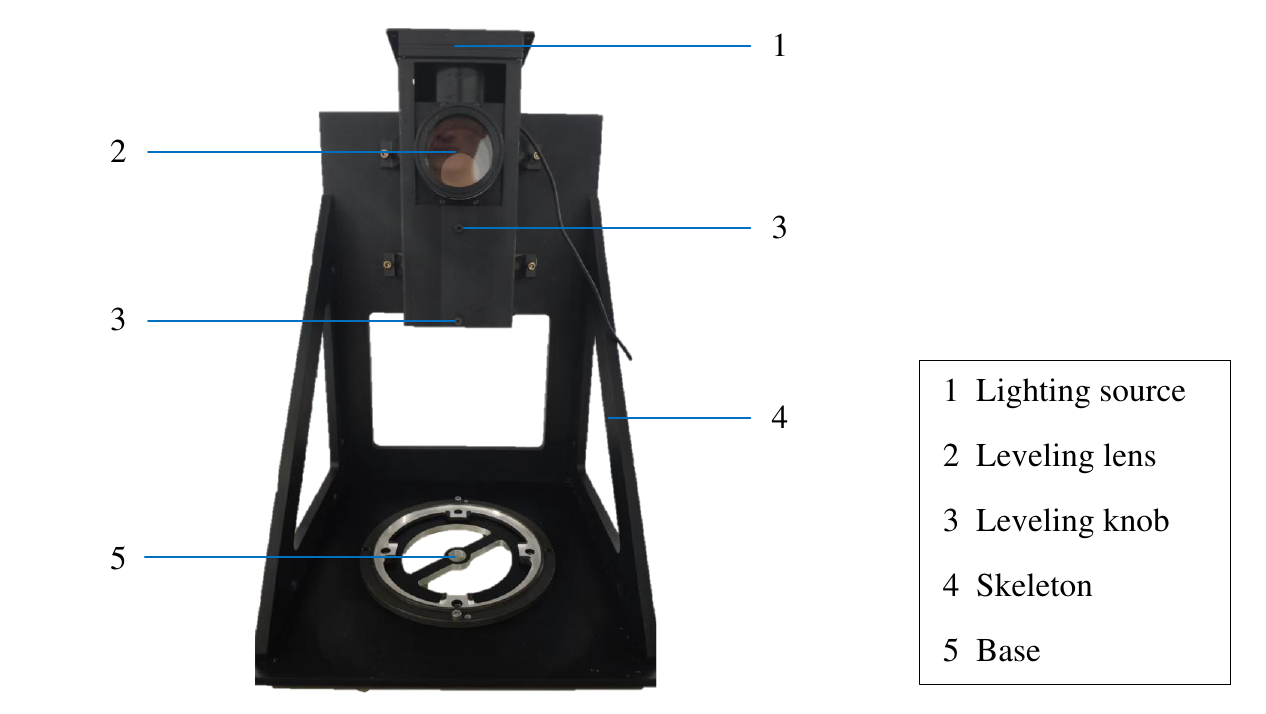} 
\caption{Reference calibration system installation diagram.} 
\label{fig2} 
\end{figure}

This section introduces a calibration frame designed to achieve accurate calibration of visual navigation devices through the integration of a collimator and an adjustment system. As shown in Fig. \ref{fig2}, the main structure of the calibration frame consists of a skeleton and a base. The skeleton, constructed from high-strength materials, supports the weight of the visual navigation device and withstands operational loads. The base, fabricated from thermally stable materials, incorporates an isolation design to minimize the impact of vibrations and thermal fluctuations. The collimator is set up on the calibration frame.

The calibration frame is equipped with a high-precision adjustment system, which includes mechanical components such as leveling knobs, leveling lenses, and crosshair wires. These components enable fine-tuning of the collimator's position to achieve physical alignment of the calibration frame. Specifically, the reticle of the collimator is aligned parallel to the base. This alignment ensures parallelism between the target coordinate system and the reference coordinate system. To meet this geometric requirement, the collimator undergoes precise machining, with leveling knobs and a leveling lens installed on its surface and a crosshair wire embedded internally. By adjusting the leveling knob and observing through the leveling lens, when the crosshair wire is aligned with the center of the target pattern, the parallelism condition between the two coordinate systems is satisfied.

\begin{figure}[htbp] 
\centering 
\includegraphics[width=0.9\textwidth]{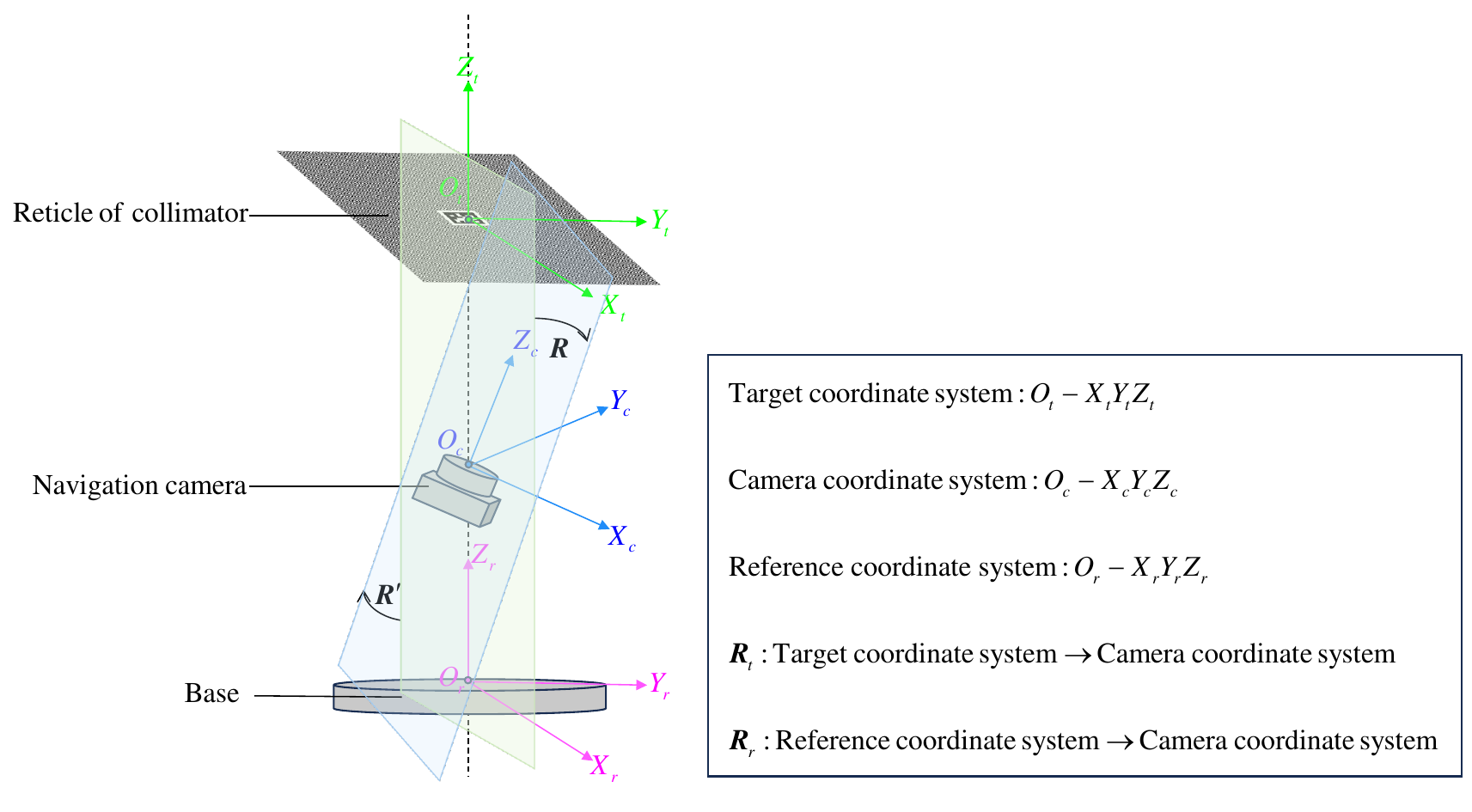} 
\caption{Coordinate system conversion schematics.} 
\label{fig3} 
\end{figure}

When conducting calibration tasks for the navigation devices, the navigation device is first mounted on the base of the calibration frame, ensuring that the reference surface of the navigation device aligns with the base surface of the calibration frame. This alignment establishes a shared reference coordinate system between the navigation device and the calibration frame. Based on the parallel relationship, the rotation matrix $\mathbf{R}_t$ of the camera relative to the collimator target can be transformed into the rotation matrix $\mathbf{R}_r$ of the camera relative to the reference, that is,  $\mathbf{R}_t = \mathbf{R}_r$. This transformation enables the determination of the camera's attitude relative to the reference. The fundamental principle of this process is illustrated in Fig. \ref{fig3}.

In practical navigation device applications, the target object is typically positioned at a considerable distance from the camera. As a result, the distance between the camera and the reference has a negligible impact on the calculation of the target's direction. Conversely, the relative rotation between the camera coordinate system and the reference coordinate system plays a critical role in determining the target's direction. Therefore, the attitude calibration task in this study primarily focuses on the calibration of the relative rotation.

\section{Principles of visual navigation device calibration}
\label{sec3}

\begin{figure}[htbp] 
\centering 
\includegraphics[width=1\textwidth]{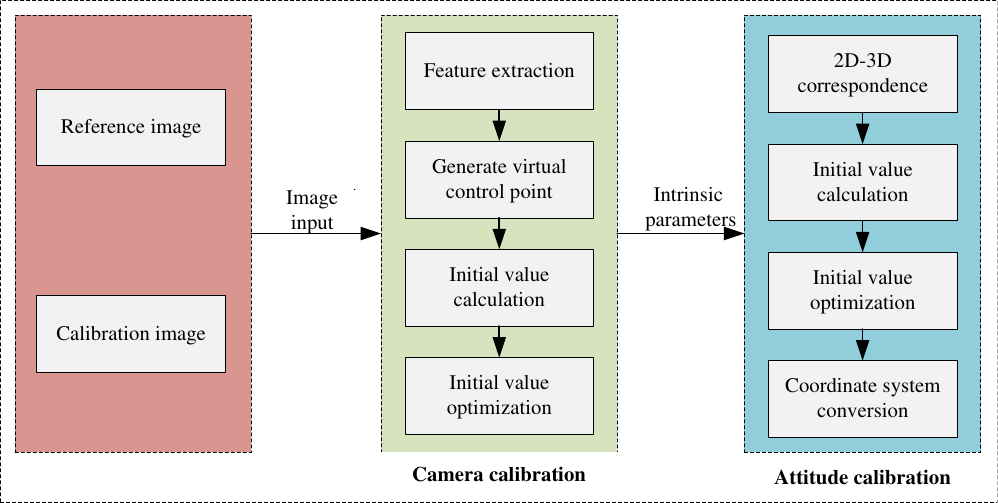} 
\caption{Visual navigation device calibration process flowchart.} 
\label{fig4} 
\end{figure}

As shown in Fig. \ref{fig4}, the calibration procedure for achieving visual navigation devices is divided into two critical segments: camera calibration and attitude calibration. Two kinds of images need to be input. One is the reference image, which is captured by the camera with known parameters, and the other is the calibration image, which is captured by the navigation camera. Feature points are identified in both images for further processing.

During the camera calibration phase, the pixel coordinates of the feature points are extracted from the reference image. Then the corresponding feature directions are calculated using the known parameters. Furthermore, a random distance is assigned within a defined range to generate virtual control points. Similarly, extracting the pixel coordinates of the feature points from the calibration image. Camera calibration can be performed through the correspondence between the pixel coordinates of the feature points and the spatial coordinates of the virtual control points.

During the attitude calibration phase, the positioning information from the Apriltag pattern \citep{Edwin2011} is used to determine three-dimensional coordinates of the encoded marker pattern. By integrating the camera parameters, two-dimensional pixel coordinates, and three-dimensional coordinates of the feature points, the rotation matrix of the camera coordinate system relative to the target coordinate system can be calculated. Finally, attitude calibration is accomplished by taking advantage of the geometric construction of the visual navigation device calibration system.

\subsection{Camera calibration}
\label{subsec31}

\begin{figure}[htbp] 
\centering 
\includegraphics[width=0.6\textwidth]{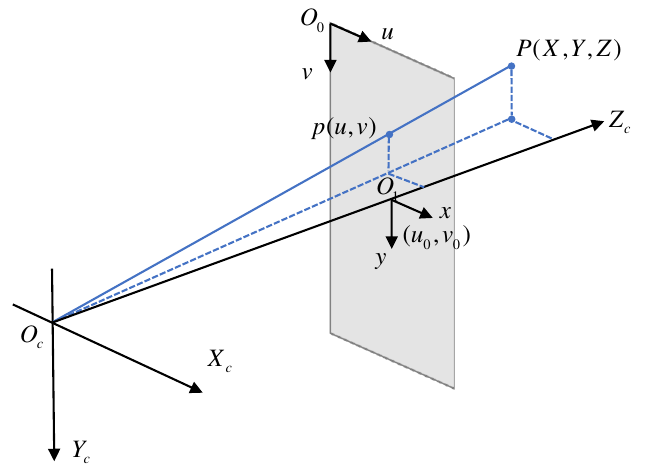} 
\caption{Central projection model.} 
\label{fig5} 
\end{figure}

As shown in Fig. \ref{fig5}, we use the central projection model to describe the camera model.  $O_{0}-uv$ represents the pixel coordinate system, $O_{1}-xy$ represents the image coordinate system, and $O_{c}-X_{c}Y_{c}Z_{c}$ represents the camera coordinate system. The three-dimensional control point $\mathbf{P}$ is projected onto the image plane of the camera, resulting in the image point $\mathbf{p}$. 

The relationship between the pixel coordinates of the image point and the spatial coordinates of the control point is given by:

\begin{equation}
\begin{cases}
\dfrac{u-u_0}{f_x}=\dfrac{X}{Z}\\
\dfrac{v-v_0}{f_y}=\dfrac{Y}{Z}
\label{eq1}
\end{cases}
\end{equation}

\noindent where $f_{x}$ and $f_{y}$ represent the focal lengths of the camera and the coordinate $(u_{0},v_{0})$ refers to the principal point. 

Divide the camera parameters into intrinsic parameters and distortion coefficients. To compensate for the distortion effects during the imaging process, we used the following distortion model \citep{Weng1992}:

\begin{equation}
\begin{cases}
\tilde{u}=u+(u-u_{0})[k_{1}(x^2+y^2)+k_{2}(x^2+y^2)^2]\\
\tilde{v}=v+(v-v_{0})[k_{1}(x^2+y^2)+k_{2}(x^2+y^2)^2]
\label{eq2}
\end{cases}
\end{equation}

\noindent where $(u,v)$ represents the actual pixel coordinates, $(\tilde{u},\tilde{v})$ denotes the ideal pixel coordinates, $k_{1}$ and $k_{2}$  are the distortion coefficients, and  $(x,y)$  denotes  normalized image coordinates.

\subsubsection{Generate virtual control points}

\begin{figure}[t] 
\centering 
\includegraphics[width=1\textwidth]{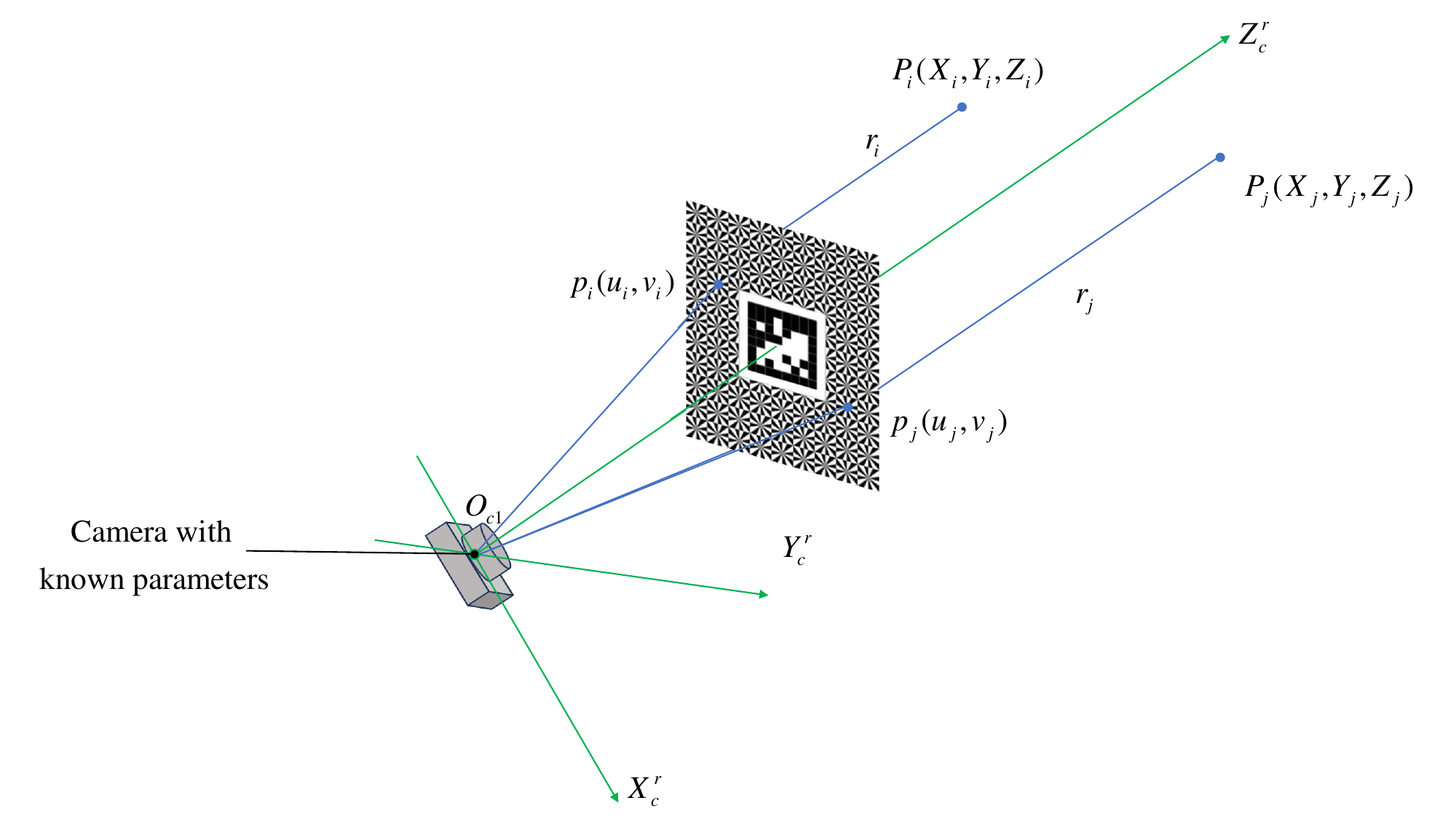} 
\caption{Generate virtual control point schematics.} 
\label{fig6} 
\end{figure}

To generate virtual control points, we use a camera with known parameters to capture an image of the reticle, which serves as the reference image. Fig. \ref{fig6} illustrates the main principle of generating virtual control points. $O_{c}-X_{c}^{r}Y_{c}^{r}Z_{c}^{r}$ represents the camera coordinate system, with the optical center of the reference camera as the origin. And $(u_{i}^{r}, v_{i}^{r})$ represents the pixel coordinates of the i-th feature point in the encoded marker pattern. $({\tilde u}_{i}^{r}, {\tilde v}_{i}^{r})$ represents the ideal image point coordinates after dedistortion processing. The corresponding feature directions are calculated as follows:

\begin{equation}
\left[ {\begin{array}{*{20}{c}}
{{x^{r}_{i}}}\\
{{y^{r}_{i}}}\\
1
\end{array}} \right] = {\left[ {\begin{array}{*{20}{c}}
{{f^{r}_{x}}}&0&{{u_{r}}}\\
0&{{f^{r}_{y}}}&{{v_{r}}}\\
0&0&1
\end{array}} \right]^{ - 1}}\left[ {\begin{array}{*{20}{c}}
{{{\tilde u}^{r}_{i}}}\\
{{{\tilde v}^{r}_{i}}}\\
1
\end{array}} \right]
\label{eq3}
\end{equation}

\noindent $f^{r}_{x}$ and $f^{r}_{y}$ denote the focal lengths of the camera with known parameters, and the principal point coordinates of the reference camera are represented by $(u_{r},v_{r})$. The three-dimensional coordinates $(X_{i}, Y_{i}, Z_{i})$ of the virtual control points are derived by scaling the feature direction vector of the feature points with a randomly assigned distance $r_{i}$, that is:

\begin{equation}
\left[ {\begin{array}{*{20}{c}}
{{X_i}}\\
{{Y_i}}\\
{{Z_i}}
\end{array}} \right] = {r_i} \left[ {\begin{array}{*{20}{c}}
{{x^{r}_{i}}}\\
{{y^{r}_{i}}}\\
1
\end{array}} \right]
\label{eq4}
\end{equation}

To simulate the operational distance of UAVs, we set the range of the random distance $r_{i}$ to be from 100 to 1000 m. Thus, a set of three-dimensional virtual control points is generated, which are uniformly distributed within the field of view and have random depths. The corresponding relationship between the virtual control points and image points can assist in completing the camera calibration task.

\subsubsection{Estimation of camera parameters}

Control the navigation camera to capture an image of the reticle, which serves as the calibration image. Extract the two-dimensional pixel coordinates of the feature points in the calibration image. Because of the unique characteristics of the collimator, the feature points $(u_{i}^{c},v_{i}^{c})$ in the calibration image correspond with the virtual control points $(X_{i}, Y_{i}, Z_{i})$. To accomplish the camera calibration task, We assume that the distortion coefficients in the central region are 0. Using the central perspective projection matrix, the relationship is given by:

\begin{align}
{s_i}\left[ {\begin{array}{*{20}{c}}
{{u^{c}_{i}}}\\
{{v^{c}_{i}}}\\
1
\end{array}} \right] &= \left[ {\begin{array}{*{20}{c}}
{{f^{c}_{x}}{r_o} + {u_c}{r_6}}&{{f^{c}_{x}}{r_1} + {u_c}{r_7}}&{{f^{c}_{x}}{r_2} + {u_c}{r_8}}&{{f^{c}_{x}}{t_x} + {u_c}{t_z}}\\
{{f^{c}_{y}}{r_3} + {v_c}{r_6}}&{{f^{c}_{y}}{r_4} + {v_c}{r_7}}&{{f^{c}_{y}}{r_5} + {v_c}{r_8}}&{{f^{c}_{y}}{t_y} + {v_c}{t_z}}\\
{{r_6}}&{{r_7}}&{{r_8}}&{{t_z}}
\end{array}} \right]\left[ {\begin{array}{*{20}{c}}
{{X_i}}\\
{{Y_i}}\\
{{Z_i}}\\
1
\end{array}} \right] \nonumber \\
&= \left[ {\begin{array}{*{20}{c}}
{{m_0}}&{{m_1}}&{{m_2}}&{{m_3}}\\
{{m_4}}&{{m_5}}&{{m_6}}&{{m_7}}\\
{{m_8}}&{{m_9}}&{{m_{10}}}&{{m_{11}}}
\end{array}} \right]\left[ {\begin{array}{*{20}{c}}
{{X_i}}\\
{{Y_i}}\\
{{Z_i}}\\
1
\end{array}} \right]
\label{eq5}
\end{align}

\noindent where $s_{i}$ is the scale factor, $f^{c}_{x}$ and  $f^{c}_{y}$ are the focal lengths of the navigation camera, $(u_{c},v_{c})$ is the principal point, $r_{0}-r_{8}$ are the elements of the rotation matrix, and ${\left[ {\begin{array}{*{20}{c}}{{t_x}}&{{t_y}}&{{t_z}}\end{array}} \right]^T}$ represents the translation vector. By organizing Eq. \eqref{eq5} into the collinearity equation, we obtain:

\begin{equation}
\left\{
\begin{array}{*{20}{c}}
{u^{c}_{i}} = \dfrac{m{}_0{X_i} + m{}_1{Y_i} + m{}_2{Z_i} + {m_3}}{m{}_8{X_i} + m{}_9{Y_i} + m{}_{10}{Z_i} + {m_{11}}} \\[10pt]
{v^{c}_{i}} = \dfrac{m{}_4{X_i} + m{}_5{Y_i} + m{}_6{Z_i} + {m_7}}{m{}_8{X_i} + m{}_9{Y_i} + m{}_{10}{Z_i} + {m_{11}}}
\end{array}
\right.
\label{eq6}
\end{equation}

Subject Eq. \eqref{eq6} to further manipulation:

\begin{equation}
\begin{cases}
{X_i}{m_0} + {Y_i}{m_1} + {Z_i}{m_2} + {m_3} - {u^{c}_{i}}{X_i}{m_8} - {u^{c}_{i}}{Y_i}{m_9} - {u^{c}_{i}}{Z_i}{m_{10}} - {u^{c}_{i}}{m_{11}} = 0 \\
{X_i}{m_4} + {Y_i}{m_5} + {Z_i}{m_6} + {m_7} - {v^{c}_{i}}{X_i}{m_8} - {v^{c}_{i}}{Y_i}{m_9} - {v^{c}_{i}}{Z_i}{m_{10}} - {v^{c}_{i}}{m_{11}} = 0
\label{eq7}
\end{cases}
\end{equation}

\noindent thus, a set of homogeneous linear equations is obtained. Given ${t_z} = {m_{11}} > 0$, we divide both sides of the equation by $m_{11}$, that is:

\begin{equation}
\begin{cases}
{X_i}{s_0} + {Y_i}{s_1} + {Z_i}{s_2} + {s_3} - {u^{c}_{i}}{X_i}{s_8} - {u^{c}_{i}}{Y_i}{s_9} - {u^{c}_{i}}{Z_i}{s_{10}} - {u^{c}_{i}} = 0 \\
{X_i}{s_4} + {Y_i}{s_5} + {Z_i}{s_6} + {s_7} - {v^{c}_{i}}{X_i}{s_8} - {v^{c}_{i}}{Y_i}{s_9} - {v^{c}_{i}}{Z_i}{s_{10}} - {v^{c}_{i}} = 0
\label{eq8}
\end{cases}
\end{equation}

The number of unknowns is eleven, where ${s_i} ={m_i}/{m_{11}}$. It's easy to know that $m_{11} = \sqrt {1/(s_8^2 + s_9^2 + s_{10}^2)}$,  due to the unit orthogonality of the rotation matrix. And Eq. \eqref{eq8} indicates that each control point can provide two equations.  In theory, the task can be solved using observations from six or more control points. It is known that:

\begin{equation}
{r_6} = {m_8}, \quad {r_7} = {m_9}, \quad {r_8} = {m_{10}}
\label{eq9}
\end{equation}

\noindent assuming that:

\begin{equation}
\begin{cases}
\mathbf{m}_1 = \begin{bmatrix} {m_0} & {m_1} & {m_2} \end{bmatrix} \\[10pt]
\mathbf{m}_2 = \begin{bmatrix} {m_4} & {m_5} & {m_6} \end{bmatrix} \\[10pt]
\mathbf{m}_3 = \begin{bmatrix} {m_8} & {m_9} & {m_{10}} \end{bmatrix}
\label{eq10}
\end{cases}
\end{equation}

\begin{equation}
\begin{cases}
\mathbf{r}_1 = \begin{bmatrix} r_0 & r_1 & r_2 \end{bmatrix} \\[10pt]
\mathbf{r}_2 = \begin{bmatrix} r_3 & r_4 & r_5 \end{bmatrix} \\[10pt]
\mathbf{r}_3 = \begin{bmatrix} r_6 & r_7 & r_8 \end{bmatrix}
\label{eq11}
\end{cases}
\end{equation}

Leveraging the unit orthogonality of the rotation matrix, we can calculate the initial values of the camera parameters:

\begin{equation}
\begin{cases}
\mathbf{m}_1 \cdot \mathbf{m}_3 = f^{c}_{x} \mathbf{r}_1 \cdot \mathbf{r}_3 + u_2 \mathbf{r}_3 \cdot \mathbf{r}_3 = u_2 \\
\mathbf{m}_2 \cdot \mathbf{m}_3 = f^{c}_{y} \mathbf{r}_2 \cdot \mathbf{r}_3 + v_2 \mathbf{r}_3 \cdot \mathbf{r}_3 = v_2 \\
\|\mathbf{m}_1 \times \mathbf{m}_3\| = f^{c}_{x} \|\mathbf{r}_1 \times \mathbf{r}_3\| + u_2 \|\mathbf{r}_3 \times \mathbf{r}_3\| = f^{c}_{x} \\
\|\mathbf{m}_2 \times \mathbf{m}_3\| = f^{c}_{y} \|\mathbf{r}_2 \times \mathbf{r}_3\| + v_2 \|\mathbf{r}_3 \times \mathbf{r}_3\| = f^{c}_{y}
\label{eq12}
\end{cases}
\end{equation}

\noindent from Eq. \eqref{eq5} and Eq. \eqref{eq12}, the initial values of the absolute extrinsic parameters can be determined as follows:

\begin{equation}
\begin{cases}
{r_0} = \left( {{m_0} - {u_2}{r_6}} \right)/{f^{c}_{x}}, \quad {r_1} = \left( {{m_1} - {u_2}{r_7}} \right)/{f^{c}_{x}}, \quad {r_2} = \left( {{m_2} - {u_2}{r_8}} \right)/{f^{c}_{x}} \\
{r_3} = \left( {{m_4} - {v_2}{r_6}} \right)/{f^{c}_{y}}, \quad {r_4} = \left( {{m_5} - {v_2}{r_7}} \right)/{f^{c}_{y}}, \quad {r_5} = \left( {{m_6} - {v_2}{r_8}} \right)/{f^{c}_{y}} \\
{t_x} = \left( {{m_3} - {u_2}{t_z}} \right)/{f^{c}_{x}}, \quad {t_y} = \left( {{m_7} - {v_2}{t_z}} \right)/{f^{c}_{y}}
\label{eq13}
\end{cases}
\end{equation}

By incorporating feature points from image edges, the initial values of the distortion coefficients $k_1$  and  $k_2$  are obtained using Eq.  \eqref{eq2}. Consequently, the initial values of the camera parameters are determined. These initial values are subsequently refined through iterative optimization to enhance calibration accuracy.

\subsubsection{Optimize the camera parameters}

To further enhance the accuracy of the parameter estimation, the least squares method is adopted as the optimization criterion, with the minimization of the re-projection error serving as the constraint condition. Specifically, the bundle adjustment method is applied to adjust the initial estimates of the camera parameters, thereby solving for accurate and robust camera parameters. The optimization objective function is:

\begin{equation}
\mathop {\min }\limits_{\mathbf{K},\mathbf{d}} \sum\limits_{i = 1}^n {\rho (\parallel \mathbf{p}_i - \pi (} \mathbf{K},\mathbf{d},\mathbf{R}_v,\mathbf{t}_v,\mathbf{P}_i){\parallel ^2})
\label{eq14}
\end{equation}

\noindent where $\mathbf{p}_i (u^{c}_{i}, v^{c}_{i})$ represents the two-dimensional pixel coordinates of the i-th feature point extracted from the calibration image, and  $\mathbf{P}_i (X_i, Y_i, Z_i)$ denotes the three-dimensional coordinates of the corresponding virtual control point, with a total of $n$  feature points. The projection equation is represented by $\pi ()$, the intrinsic parameter matrix of the navigation camera is denoted by $\mathbf{K}$, and the distortion coefficients are denoted by  $\mathbf{d}$. The absolute extrinsic parameters of the target coordinate system relative to the camera coordinate system are represented by the rotation matrix  
$\mathbf{R}_v$ and the translation vector $\mathbf{t}_v$  and the cost function is denoted by $\rho ()$. 

Through continuous optimization and iteration, more precise camera parameters are ultimately obtained. In the actual optimization process, the rotation matrix $\mathbf{R}_v$ is expressed in the form of an axis-angle $\mathbf{r}_v$, which can be converted to and from the rotation matrix using the Rodrigues formula:

\begin{equation}
\mathbf{R}_{{v}} = \exp \left( \left[ \mathbf{r}_{{v}} \right]_{\times} \right) = \mathbf{I} + \left( \sin \| \mathbf{r}_{{v}} \| \right) \left[ \mathbf{\hat{r}}_{{v}} \right]_{\times} + \left( 1 - \cos \| \mathbf{r}_{{v}} \| \right) \left[ \mathbf{\hat{r}}_{{v}} \right]_{\times}^2
\label{eq15}
\end{equation}

\noindent where  $\mathbf{I}$ denotes the identity matrix, $\| \mathbf{r}_{{v}} \|$ 
 denotes the rotation angle, $\mathbf{\hat{r}}_{{v}} = \mathord{\mathbf{r}}_{{v}} / \| \mathbf{r}_{{v}} \|$ denotes the rotation axis, and $\left[ \mathbf{\hat{r}}_{{v}} \right]_{\mathrm{x}}$ denotes the skew-symmetric matrix of the rotation axis.

\subsection{Attitude calibration}
\label{subsec32}

\subsubsection{Solution for initial values of relative rotation}

After obtaining high-precision camera parameters, the extracted pixel coordinates
$(u^{c}_{i}, v^{c}_{i})$ are subjected to a de-distortion process, resulting in ideal coordinates
$({\tilde u}^{c}_{i}, {\tilde v}^{c}_{i})$. Using information from the Apriltag pattern, the coordinates $(X^{t}_{i}, Y^{t}_{i}, Z^{t}_{i})$ of the feature points are determined. It is worth mentioning that these groups of three-dimensional points lie on a common plane. 

By matching the ideal pixel coordinates $({\tilde u}^{c}_{i}, {\tilde v}^{c}_{i})$ with the three-dimensional coordinates $(X^{t}_{i}, Y^{t}_{i}, Z^{t}_{i})$, the rigid body transformation relationship between the target coordinate system and the calibration camera coordinate system can be calculated:

\begin{equation}
\left[ {\begin{array}{*{20}{c}}
{{{\tilde u}^{c}_{i}}}\\
{{{\tilde v}^{c}_{i}}}\\
1
\end{array}} \right] = {\lambda _i}\mathbf{K}\left[ {\begin{array}{*{20}{c}}
{\mathbf{R}_t}&{{\mathbf{t}_t}}
\end{array}} \right]\left[ {\begin{array}{*{20}{c}}
{{X^{t}_{i}}}\\
{{Y^{t}_{i}}}\\
{{Z^{t}_{i}}}\\
1
\end{array}} \right] = {\lambda _i}\mathbf{K}\left[ {\begin{array}{*{20}{c}}
{{\mathbf{r}^{t}_{1}}}&{{\mathbf{r}^{t}_{2}}}&{{\mathbf{r}^{t}_{3}}}&{{\mathbf{t}_t}}
\end{array}} \right]\left[ {\begin{array}{*{20}{c}}
{{X^{t}_{i}}}\\
{{Y^{t}_{i}}}\\
{{Z^{t}_{i}}}\\
1
\label{eq16}
\end{array}} \right]
\end{equation}

\noindent where ${\lambda _i}$ is the reciprocal of the scale factor, $\mathbf{K}$ is the intrinsic parameter matrix, $\mathbf{R}_{t} = \left[ {\begin{array}{*{20}{c}}
{{\mathbf{r}^{t}_{1}}}&{{\mathbf{r}^{t}_{2}}}&{{\mathbf{r}^{t}_{3}}}
\end{array}} \right] \quad \text{and} \quad \mathbf{t}_{t}$ represent the rotation matrix and translation vector.

By establishing the target coordinate system on the reticle, we have $Z^{t}_{i} = 0$, thus simplifying Eq. \eqref{eq16} to:

\begin{equation}
\left[ {\begin{array}{*{20}{c}}
{{{\tilde u}^{c}_{i}}}\\
{{{\tilde v}^{c}_{i}}}\\
1
\end{array}} \right] = {\lambda _i}\mathbf{K}\left[ {\begin{array}{*{20}{c}}
{{\mathbf{r}^{t}_{1}}}&{{\mathbf{r}^{t}_{2}}}&{{\mathbf{t}_{t}}}
\end{array}} \right]\left[ {\begin{array}{*{20}{c}}
{{X^{t}_{i}}}\\
{{Y^{t}_{i}}}\\
1
\end{array}} \right] = \mathbf{H}\left[ {\begin{array}{*{20}{c}}
{{X^{t}_{i}}}\\
{{Y^{t}_{i}}}\\
1
\label{eq17}
\end{array}} \right]
\end{equation}

\noindent where  $\mathbf{H}$ represents the homography matrix, assuming that:

\begin{equation}
\mathbf{H} = \left[ {\begin{array}{*{20}{c}}
{{\mathbf{h}_1}}&{{\mathbf{h}_2}}&{{\mathbf{h}_3}}
\label{eq18}
\end{array}} \right]
\end{equation}

Based on the unit orthogonality of the rotation matrix, the initial values of the extrinsic parameters can be determined as:

\begin{equation}
\left\{ {\begin{array}{*{20}{c}}
{{\mathbf{r}^{t}_{1}} = (\mathbf{K}^{-1} {\mathbf{h}_1}) / \| \mathbf{K}^{-1} {\mathbf{h}_1} \| }\\
{{\mathbf{r}^{t}_{2}} = (\mathbf{K}^{-1} {\mathbf{h}_2}) / \| \mathbf{K}^{-1} {\mathbf{h}_2} \| }\\
{{\mathbf{t}_{t}} = (\mathbf{K}^{-1} {\mathbf{h}_3}) / \| \mathbf{K}^{-1} {\mathbf{h}_3} \| }\\
{{\mathbf{r}^{t}_{3}} = {\mathbf{r}^{t}_{1}} \times {\mathbf{r}^{t}_{2}}}
\label{eq19}
\end{array}} \right.
\end{equation}

Similarly, the rotation matrix $\mathbf{R}_{t}$ is represented by the axis angle $\mathbf{r}_{t}$.

\subsubsection{Optimizing relative rotation}

The Levenberg-Marquardt (LM) algorithm is utilized to iteratively optimize the initial values of the rotation axis angles, with the minimization of the re-projection error serving as the objective function. Thereby we obtain a precise solution for the relative rotation. The optimization function is as follows:

\begin{equation}
\mathop {{\rm{arc}}\min }\limits_{\mathbf{r}_{t}} \sum\limits_{i = 1}^n {\| \mathbf{p}_i - \pi(\mathbf{K}, \mathbf{d}, \mathbf{r}_{t}, \mathbf{t}_{t}, \mathbf{P}^{t}_{i}) \|^2}, 
\label{eq20}
\end{equation}

\noindent where $\mathbf{r}_{t}$ and $\mathbf{t}_{t}$  represent the rotation axis angle and translation vector, respectively, from the target coordinate system to the calibration camera coordinate system. After optimization and iteration, the precise rotation axis angle is ultimately obtained. As shown in Section \ref{subsec22}, $\mathbf{R}_t = \mathbf{R}_r$, thus we obtain the rotation matrix $\mathbf{R}_r$ of the camera relative to the collimator base and complete the attitude calibration.

\section{Experimental results and analysis}
\label{sec4}

The calibration process of the visual navigation device based on the collimator mainly consists of two key parts: first, calibrating the navigation camera through a single calibration image; second, achieving attitude calibration by solving the relative rotation between the camera and the collimator reticle. To evaluate the robustness and accuracy of the proposed algorithm, comparative experiments were conducted for both segments.

\subsection{Experiment on camera calibration using a single image}
\label{subsec41}

In this experiment, the proposed camera calibration algorithm was compared with Zhang's calibration method. Two camera types were utilized: one was the reference camera, with known parameters, used to generate virtual control points; the other was the test camera, used to evaluate the performance of the algorithms.

The experimental procedure was designed as follows. First, 30 checkerboard images were captured separately by the reference and test cameras. Subsequently, the parameters of both cameras were calibrated using Zhang's calibration method \citep{Zhang2000}. Next, the reference and test cameras were controlled to capture a reticle image. Following this, virtual control points were generated based on the parameters of the reference camera and the feature direction calculated from the above image, with control points' depths ranging from 100 to 1000 m. Finally, calculate the parameters of the test camera. These results were compared with those calculated using Zhang's calibration method.

Additionally, to verify the robustness of the calibration algorithm under different conditions, three check experiments were designed.  These experiments focused on the relationship between the key parameters of the reference and test cameras, specifically the resolution and focal length:

\begin{enumerate}
    \item The reference and test cameras had the same resolution and focal length. Both the reference and test cameras had a resolution of 2448 × 2048 pixels and a focal length of 20 mm.
    \item The reference and test cameras had the same resolution but different focal lengths. Both cameras had a resolution of 2448 × 2048 pixels; the reference camera had a focal length of 20 mm, whereas the test camera had a focal length of 25 mm.
    \item Both the resolution and focal length differed between the reference and test cameras. The reference camera had a resolution of 2448 × 2048 pixels and a focal length of 20 mm, whereas the test camera had a resolution of 1280 × 1024 pixels and a focal length of 12 mm.
\end{enumerate}

\begin{figure}[htbp]
\centering
\begin{subfigure}[b]{0.23\textwidth}
    \centering
    \includegraphics[width=\textwidth]{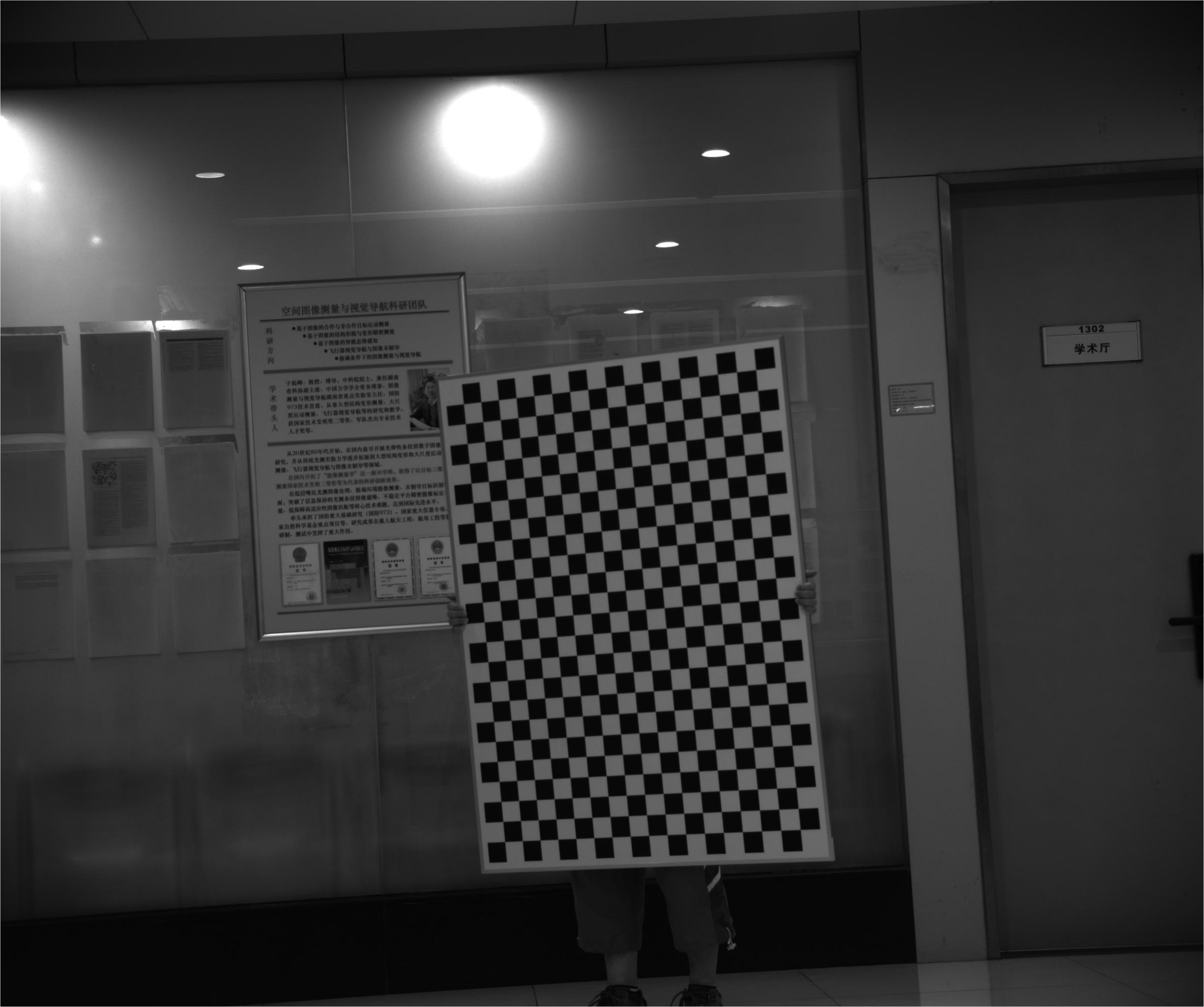}
    \caption{}
    \label{fig7a}
\end{subfigure}
\hfill
\begin{subfigure}[b]{0.23\textwidth}
    \centering
    \includegraphics[width=\textwidth]{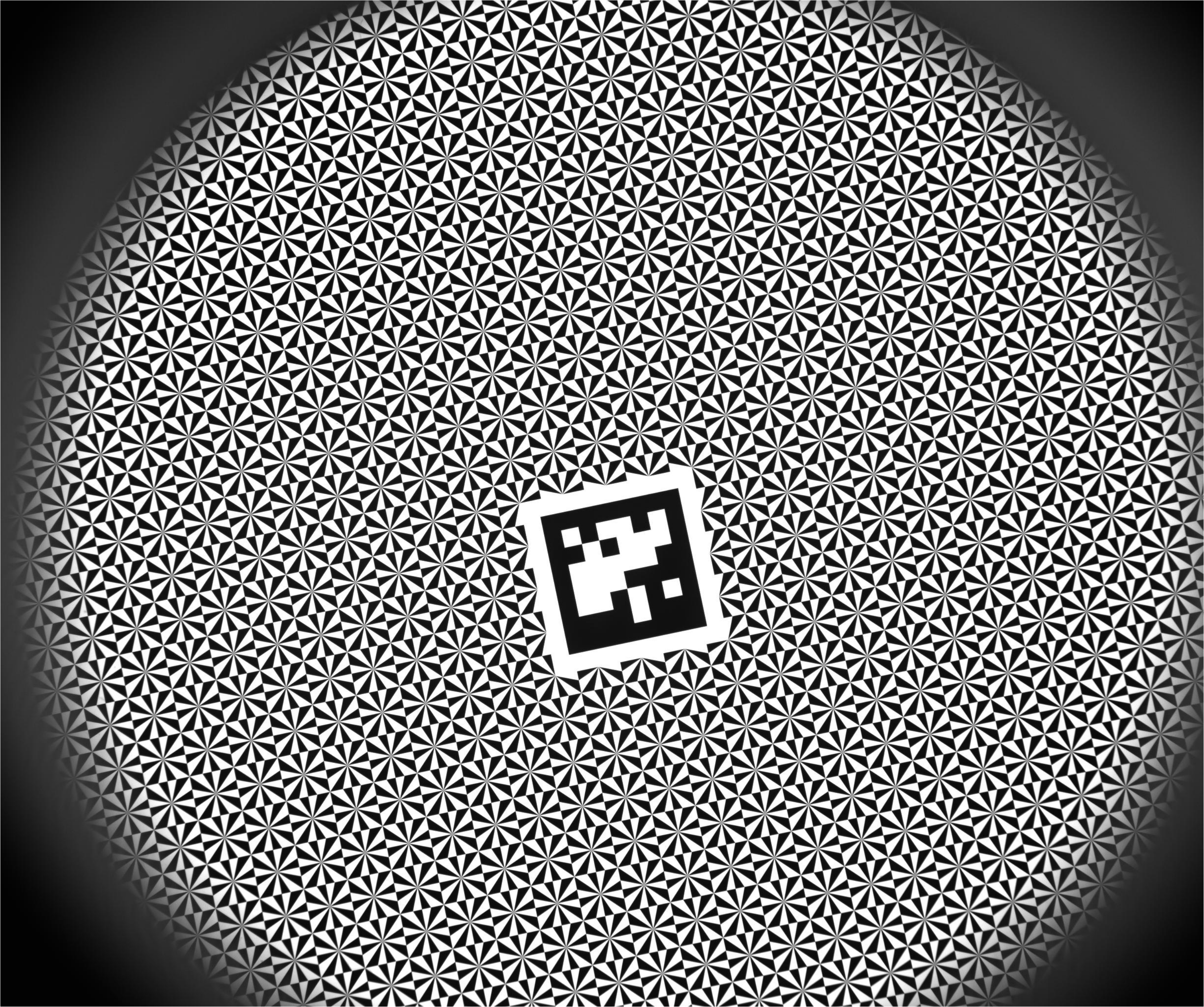}
    \caption{}
    \label{fig7b}
\end{subfigure}
\hfill
\begin{subfigure}[b]{0.23\textwidth}
    \centering
    \includegraphics[width=\textwidth]{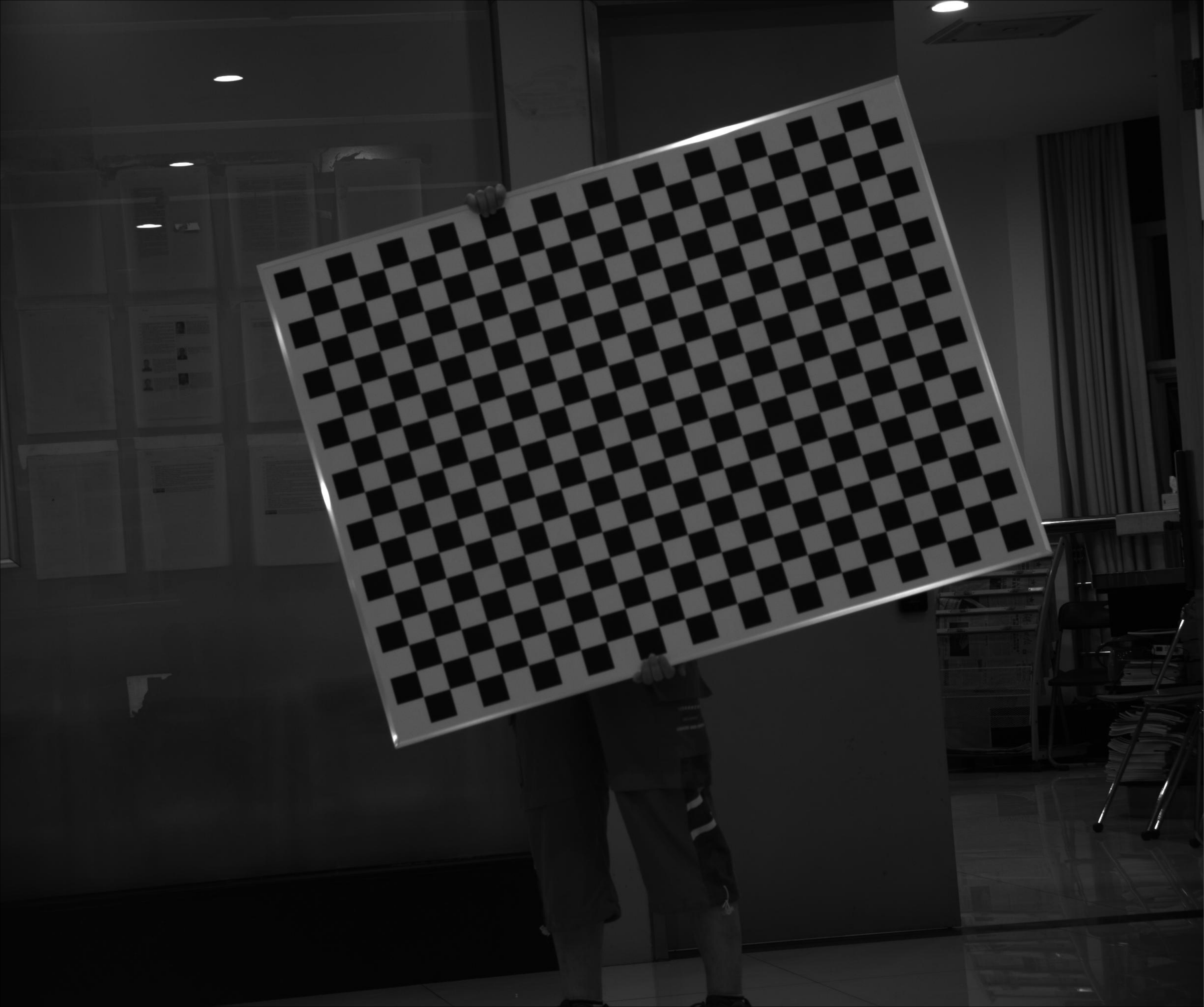}
    \caption{}
    \label{fig7c}
\end{subfigure}
\hfill
\begin{subfigure}[b]{0.23\textwidth}
    \centering
    \includegraphics[width=\textwidth]{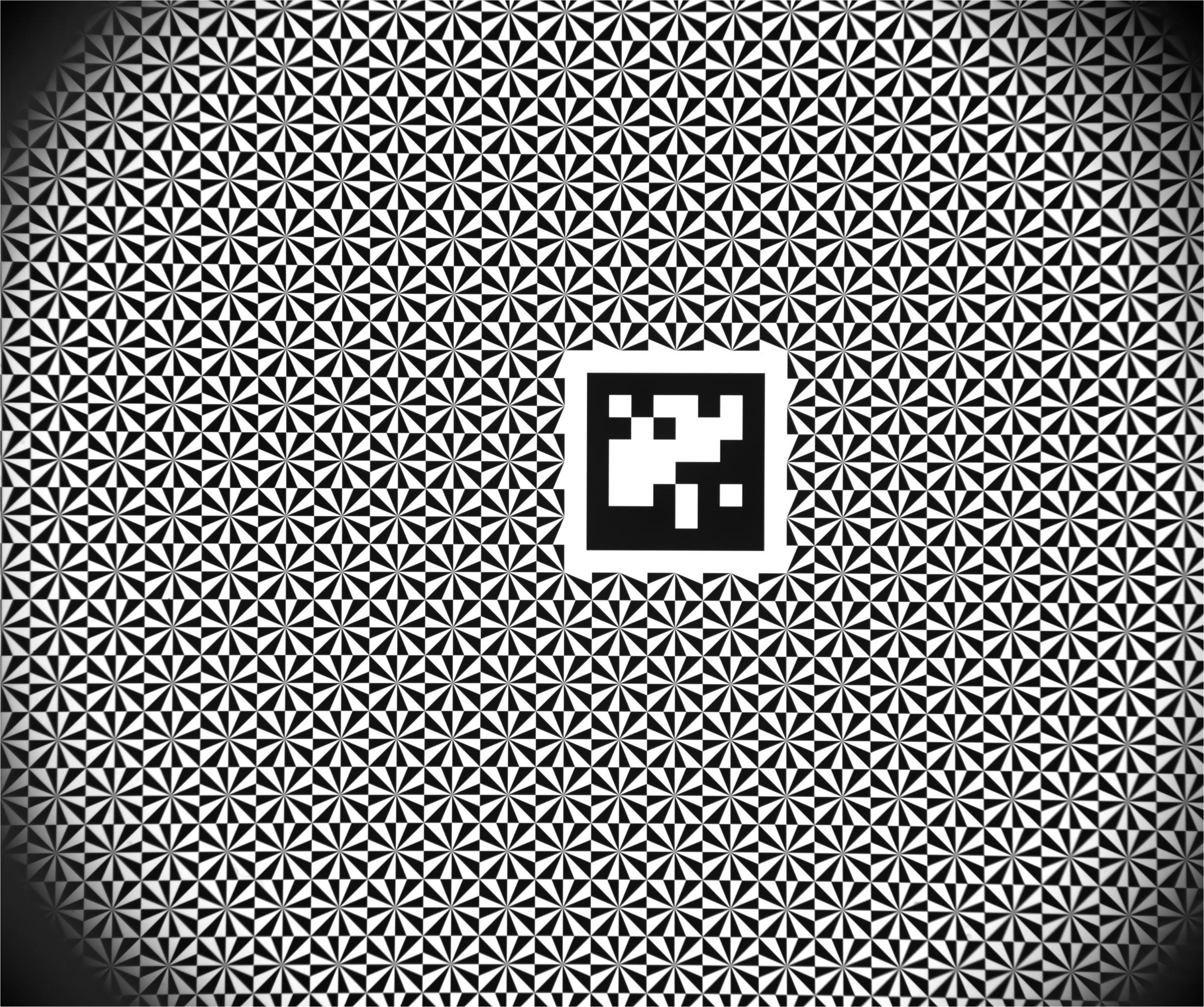}
    \caption{}
    \label{fig7d}
\end{subfigure}
\caption{Partially calibrated images at different focal lengths. (a) (b) Calibrated image at 20 mm focal length; (c) (d) Calibrated image at 25 mm focal length.}
\label{fig7}
\end{figure}

As shown in Fig. \ref{fig7}, it displays some calibration images captured under different focal lengths. Specifically, Fig. \ref{fig7a} and Fig. \ref{fig7b} represent calibration images taken at a focal length of 20 mm, while Fig. \ref{fig7c} and Fig. \ref{fig7d} represent calibration images taken at a focal length of 25 mm.

As shown in Table \ref{tab1}, the differences between the proposed calibration method in this study and Zhang's calibration method are not significant. The re-projection errors are roughly equivalent and both meet the accuracy requirements for engineering applications. However, Zhang's calibration method requires a rich set of imaging poses across the entire field of view. In contrast, our proposed method in this study only needs to capture a full-field image of the collimator reticle. In the task of industrial mass calibration of visual navigation devices, our method shows obvious advantages with higher efficiency and practicality.

\begin{table}[]
\caption{Comparison of calibration results obtained using the proposed method and Zhang's method under different conditions}
\centering
\label{tab1}
\resizebox{\textwidth}{!}{%
\begin{tabular}{cccccccccccc}
\hline
Intrinsic                & \multicolumn{3}{c}{a}    &  & \multicolumn{3}{c}{b}    &  & \multicolumn{3}{c}{c}     \\ \cline{2-4} \cline{6-8} \cline{10-12} 
parameter                & Classic & Our    & Mean  &  & Classic & Our    & Mean  &  & Classic & Our     & Mean  \\
\multicolumn{1}{l}{} &
  \multicolumn{1}{l}{method} &
  \multicolumn{1}{l}{method} &
  \multicolumn{1}{l}{error} &
  \multicolumn{1}{l}{} &
  \multicolumn{1}{l}{method} &
  \multicolumn{1}{l}{method} &
  \multicolumn{1}{l}{error} &
  \multicolumn{1}{l}{} &
  \multicolumn{1}{l}{method} &
  \multicolumn{1}{l}{method} &
  \multicolumn{1}{l}{error} \\ \hline
${f_{x}^{c}}/{\rm{pixel}}$  & 5967.7  & 5968.0 & 0.3   &  & 7521.3  & 7537.8 & 16.5  &  & 2679.5  & 2689.5  & 10.0  \\
${f_{y}^{c}}/{\rm{pixel}}$  & 5969.0  & 5967.9 & 1.1   &  & 7521.4  & 7536.8 & 15.4  &  & 2678.9  & 2688.8  & 9.9   \\
${u_{2}}/{\rm{pixel}}$   & 1222.4  & 1227.0 & 4.6   &  & 1227.8  & 1232.6 & 4.8   &  & 625.40  & 632.26  & 6.84  \\
${v_{2}}/{\rm{pixel}}$   & 1023.5  & 1014.4 & 9.1   &  & 1015.2  & 1019.4 & 4.2   &  & 514.64  & 521.33  & 6.69  \\
$k_{1}$                  & 0.2380  & 0.2414 & 0.003 &  & 0.6093  & 0.5015 & 0.108 &  & -0.2553 & -0.1956 & 0.060 \\
$k_{2}$                  & 2.0007  & 1.9126 & 0.088 &  & 0.2327  & 3.4372 & 3.205 &  & 1.5709  & 0.1550  & 1.416 \\
re-projection error/pixel & 0.1677  & 0.1463 & -     &  & 0.1207  & 0.1389 & -     &  & 0.0393  & 0.0645  & -     \\ \hline
\end{tabular}%
}
\end{table}

Furthermore, a comparison of the results from the three experiments reveals that the re-projection error does not exhibit a clear regular variation with changes in the key parameters. This phenomenon confirms the applicability and robustness of our proposed method under common experimental conditions. It should be noted that, under the condition of distortion coefficient coupling, the distortion effect produced by $k_2$  is much smaller than that of $k_1$. Consequently, the notable difference in the distortion coefficients aligns with the expected results for a correct calibration.

\subsection{Attitude calibration verification experiment}
\label{subsec42}

After determining the camera parameters, attitude calibration can be achieved through the geometric structure of the collimator. Validation experiments were also conducted using a high-precision IMU and GPS. Because the data provided by the IMU and GPS are of high precision, the direction calculated based on these data can be considered as the true value.

The attitude calibration verification experiment was conducted as follows. First, a high-precision IMU was mounted on the reference surface of the navigation device to capture its attitude changes. Meanwhile, a GPS module was installed on the reference surface to acquire real-time positioning data. Then, a cooperative target was placed on the ground, and the navigation device was installed on a UAV for a flight test. Through RTK technology, the cooperative target was precisely geo-located to obtain its exact coordinates in the geographical space, with the UAV's flight altitude set at 1.5 km during this experiment.

\begin{figure}[htbp]
\centering
\begin{subfigure}[b]{0.46\textwidth}
    \centering
    \includegraphics[width=\textwidth]{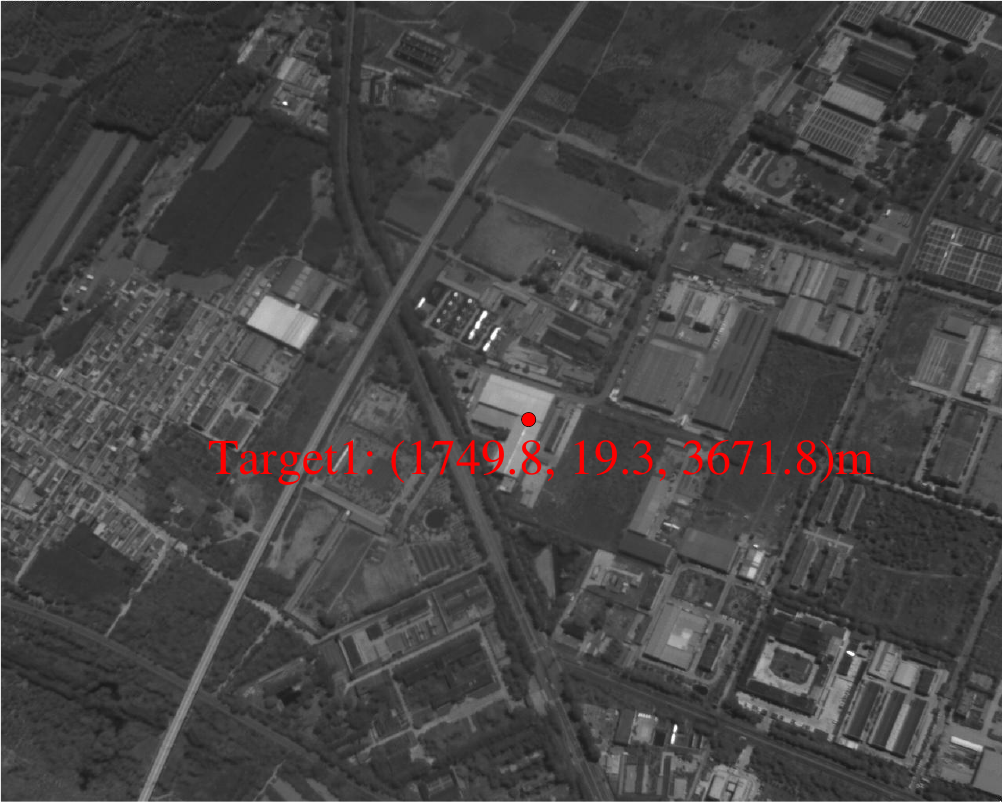}
    \caption{}
    \label{fig8a}
\end{subfigure}
\hfill
\begin{subfigure}[b]{0.49\textwidth}
    \centering
    \includegraphics[width=\textwidth]{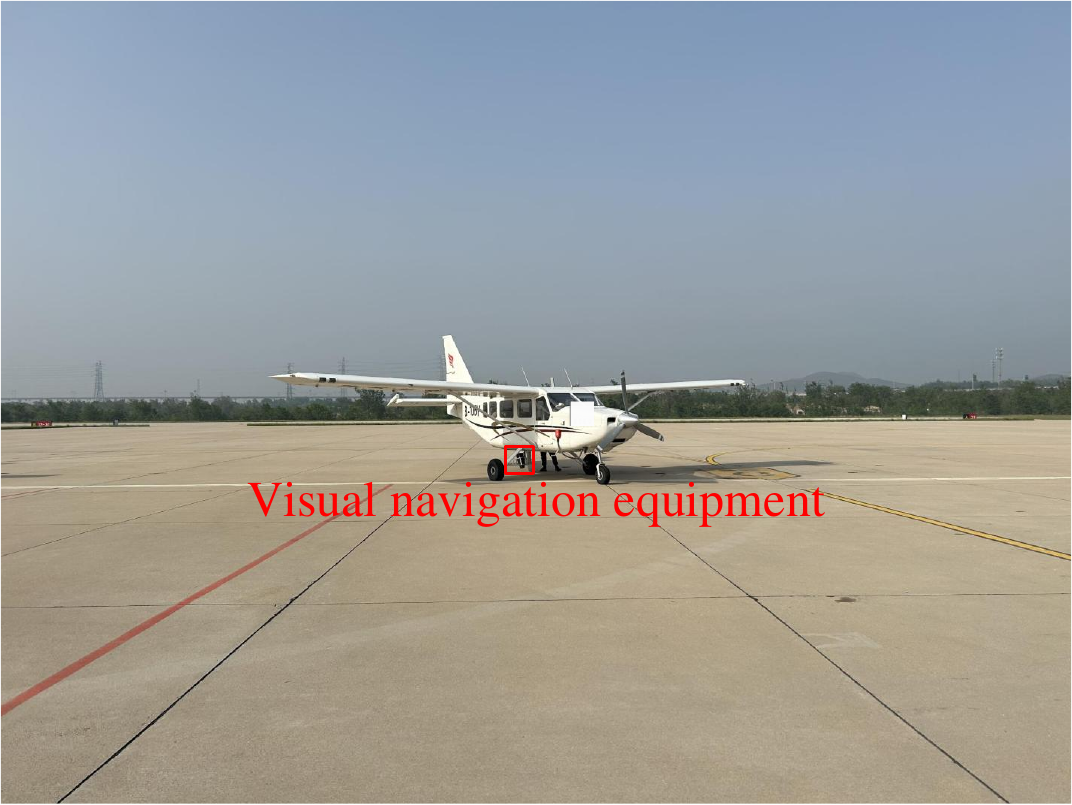}
    \caption{}
    \label{fig8b}
\end{subfigure}
\caption{Images collected from the experiment. (a) A single experimental image; (b) experimental site image.}
\label{fig8}
\end{figure}

Furthermore, using the known spatial position of the target, position, and attitude information of the navigation device, the direction vector of the target relative to the navigation device's coordinate system was calculated. Following this, the navigation camera was used to image the target. The captured image is shown in Fig. \ref{fig8}, where Fig. \ref{fig8a} is a single experimental image collected by the UAV, and Fig. \ref{fig8b} is the image of the experimental site. The pixel coordinates of the target in the image were extracted using the proposed algorithm. Based on these pixel coordinates, the camera's intrinsic parameters, and the attitude calibration results, the direction of the target in the coordinate system was computed. Finally, the direction vectors obtained from the two methods were compared to evaluate the accuracy of the attitude calibration algorithm.

We first completed the camera calibration in the laboratory. The calibration results are presented in Table \ref{tab2}. To improve readability, the relative rotation matrix is converted into Euler angles for the three principal axes. Analysis of the calibration results reveals that the angle between the camera coordinate system and the reference coordinate system is relatively small. However, given the long observation distance of the navigation device, even minor angular deviations can lead to significant navigation and positioning errors. Therefore, this error must be carefully considered in the analysis.

\begin{table}[]
\caption{Reference calibration outcomes}
\centering
\label{tab2}
\resizebox{\textwidth}{!}{%
\begin{tabular}{cccccccccc}
\hline
\multicolumn{6}{c}{Intrinsic parameter} &
   &
  \multicolumn{3}{c}{orientation parameters} \\ \cline{1-6} \cline{8-10} 
${f_{cx}}/{\rm{pixel}}$ &
  ${f_{cy}}/{\rm{pixel}}$ &
  ${C_{cx}}/{\rm{pixel}}$ &
  ${C_{cy}}/{\rm{pixel}}$ &
  $k_1$ &
  $k_2$ &
   &
  ${\theta _x}/^\circ $ &
  ${\theta _y}/^\circ $ &
  ${\theta _z}/^\circ $ \\
2677.9 &
  2678.5 &
  634.66 &
  524.12 &
  -0.2011 &
  0.1989 &
   &
  -1.5324 &
  -0.0632 &
  -0.4851 \\ \hline
\end{tabular}%
}
\end{table}

To validate the accuracy of the proposed algorithm, initial parameters were set for comparative analysis. The initial focal length of the camera was calculated as the equivalent focal length divided by the pixel size, yielding$12/4.5 \times 1000{\rm{pixel}} = 2666.7{\rm{pixel}}$. The principal point was set at the image center, that is, (640, 650) pixel, the distortion coefficients are set to zero, that is,  ${k_1} = {k_2} = 0$, while the distortion coefficients were initialized to zero, that is,  ${\theta _x} = {\theta _y} = {\theta _z} = 0^\circ $.

At the experimental site, a UAV was deployed to conduct tests at three distinct target positions within the same local coordinate system. The coordinates of the targets were as follows: Target 1 at (1749.8, 19.3, 3671.8) m, Target 2 at (4679.4, -45.3, -462.4) m, and Target 3 at (3263.9, -37.1, 955.2) m. For clarity, the unit target direction vectors were converted into yaw and pitch angles in the reference coordinate system. Three control groups were established for comparison: (1) the reference values derived from IMU and GPS data, (2) the initial values calculated using the initial parameters, and (3) the computational values obtained from the proposed algorithm.

\begin{figure}[t]
\centering
\begin{subfigure}[b]{1\textwidth}
    \centering
    \includegraphics[width=\textwidth]{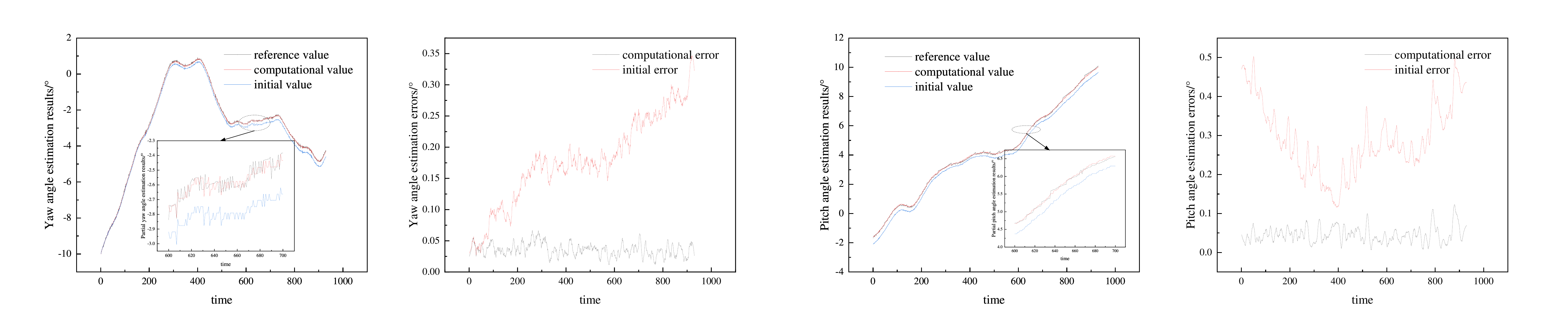}
    \caption{}
    \label{fig9a}
\end{subfigure}
\hfill
\begin{subfigure}[b]{1\textwidth}
    \centering
    \includegraphics[width=\textwidth]{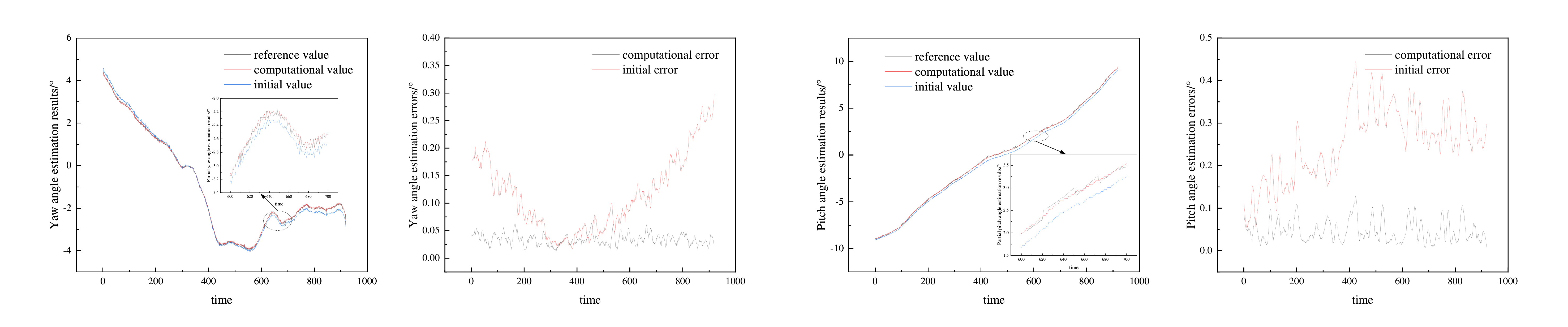}
    \caption{}
    \label{fig9b}
\end{subfigure}

\begin{subfigure}[b]{1\textwidth}
    \centering
    \includegraphics[width=\textwidth]{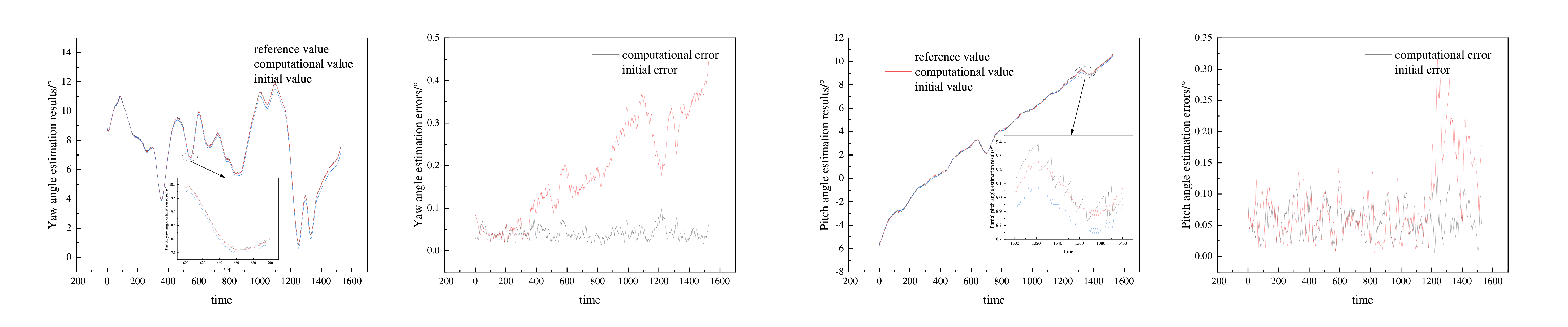}
    \caption{}
    \label{fig9c}
\end{subfigure}
\caption{Results of attitude calibration comparative experiment. (a) target1: (1749.8, 19.3, 3671.8) m; (b) target2: (4679.4, -45.3, -462.4) m; (c) target3: (3263.9, -37.1, 955.2) m.
}
\label{fig9}
\end{figure}

As shown in Fig. \ref{fig9},  the computed yaw and pitch angles for the unit target in the reference coordinate system exhibit significantly higher consistency with the values obtained from high-precision IMU and GPS data compared to those derived using the initial parameters. This demonstrates the improved accuracy achieved through the proposed calibration method. 

Taking the measurement results of Target 1 as an example, the difference between the yaw angle calculated using the initial parameters and the reference value had an average of 0.1790° and a standard deviation of 0.0706°. After recalibration with the refined parameters, the average difference between the yaw angle and the reference value was reduced to 0.0358°, representing an 80.0\% reduction, while the standard deviation dropped to 0.0102°. Similarly, the average difference for the pitch angle calculated using the initial parameters was 0.2912°, with a standard deviation of 0.0898°. With the re-calibrated parameters, the average difference decreased to 0.0457°, an 84.3\% reduction, and the standard deviation was reduced to 0.0202°. These results underscore the critical importance of precise calibration for navigation devices and demonstrate the robustness and accuracy of the proposed algorithm.

\section{Conclusion}
\label{sec5}

We propose a calibration system and method for visual navigation devices based on the collimator. This approach eliminates the traditional need for multi-angle and full-field image acquisition of the target, enabling high-precision camera calibration from a single calibration image. Additionally, the method does not rely on high-precision turntables or angle-measuring instruments. Instead, it employs a simple calibration frame with specific geometric features in conjunction with a collimator to achieve efficient attitude calibration. The system incorporates an encoded marker pattern, which includes a star-shaped pattern and an Apriltag pattern. The rich gradient information of the star-shaped pattern significantly enhances the accuracy and efficiency of feature extraction, while the Apriltag pattern ensures precise positioning and matching of control points, guaranteeing the accuracy of the calibration process.

The precision and robustness of the proposed method are validated through comparative experiments. In the camera calibration experiment, the camera parameters obtained using our method show high consistency with those calibrated from 30 images using Zhang's method. The maximum re-projection error is 0.14626 pixels, demonstrating high accuracy across various usage scenarios and meeting the requirements of engineering applications. In the attitude calibration experiment, the unit target direction vectors, estimated by the calibrated attitude of the proposed method in terms of yaw and pitch angles in the reference coordinate system, exhibit high consistency with the results derived from high-precision IMU and GPS data. The maximum average difference is 0.0586°, and the maximum standard deviation is 0.0257°, indicating high precision and robustness. These experiments confirm that the proposed calibration system and method for visual navigation devices have significant practical application value.

\appendix

\section*{Acknowledgment}

This work is supported by Hunan Provincial Natural Science Foundation for Excellent Young Scholars (Grant 2023J120045) and National Natural Science Foundation of China (Grant 12372189).

\bibliographystyle{elsarticle-num-names} 
\bibliography{references.bib}

\end{document}